\documentclass[10pt,twocolumn,letterpaper]{article}

\usepackage[pagenumbers]{iccv}      

\usepackage{multirow}

\definecolor{iccvblue}{rgb}{0.21,0.49,0.74}
\usepackage[pagebackref,breaklinks,colorlinks,allcolors=iccvblue]{hyperref}

\def\paperID{10233} 
\def\confName{ICCV}
\def\confYear{2025}

\title{Perceive, Understand and Restore: Real-World Image Super-Resolution with Autoregressive Multimodal Generative Models}

\author{Hongyang Wei$^{1,3,\ast}$,  Shuaizheng Liu$^{2,3,\ast}$,  Chun Yuan$^{1,\dagger}$,  Lei Zhang$^{2,3,\dagger}$ \\
{$^{1}$Tsinghua Shenzhen International Graduate School, Tsinghua University} \\
{$^{2}$The Hong Kong Polytechnic University \qquad $^{3}$OPPO Research Institute}
 \\
{\tt\small \ weihy23@mails.tsinghua.edu.cn, \ shuaizhengliu21@gmail.com} \  \\
{\tt\small \ yuanc@sz.tsinghua.edu.cn, \ cslzhang@comp.polyu.edu.hk}
\\
}

\begin{document}
\maketitle
\begin{abstract}

By leveraging the generative priors from pre-trained text-to-image diffusion models, significant progress has been made in real-world image super-resolution (Real-ISR). However, these methods tend to generate inaccurate and unnatural reconstructions in complex and/or heavily degraded scenes, primarily due to their limited perception and understanding capability of the input low-quality image. 
To address these limitations, we propose, for the first time to our knowledge, to adapt the pre-trained autoregressive multimodal model such as Lumina-mGPT into a robust Real-ISR model, namely \textbf{PURE}, which \textbf{P}erceives and \textbf{U}nderstands the input low-quality image, then \textbf{RE}stores its high-quality counterpart. 
Specifically, we implement instruction tuning on Lumina-mGPT to perceive the image degradation level and the relationships between previously generated image tokens and the next token, understand the image content by generating image semantic descriptions, and consequently restore the image by generating high-quality image tokens autoregressively with the collected information. In addition, we reveal that the image token entropy reflects the image structure and present a entropy-based Top-$k$ sampling strategy to optimize the local structure of the image during inference. Experimental results demonstrate that PURE preserves image content while generating realistic details, especially in complex scenes with multiple objects, showcasing the potential of autoregressive multimodal generative models for robust Real-ISR. The model and code will be available at \url{https://github.com/nonwhy/PURE}.

\end{abstract}

\renewcommand{\thefootnote}{} 
\footnotetext{$\ast$ Equal Contribution.}
\footnotetext{$\dagger$ Corresponding Authors. This work is supported by the PolyU-OPPO Joint Innovative Research Center.}    
\section{Introduction}
\label{sec:intro}
Real-world image super-resolution (Real-ISR) is a challenging task in computer vision, aiming to reconstruct high-quality (HQ) images from low-quality (LQ) inputs degraded by complex combinations of factors such as blur, noise, downsampling, \etc. Traditional discriminative image restoration methods \cite{srcnn, edsr, rcan, swinir, hat} are typically designed for optimizing image fidelity using $L_1$ or $L_2$ losses, tending to produce over-smoothed results. While generative adversarial network (GAN \cite{gan})-based approaches \cite{srgan, realesrgan} can generate more visually appealing details, they suffer from unstable training and unnatural visual artifacts.

The recently developed large-scale pre-trained text-to-image (T2I) generation models, such as Stable Diffusion (SD) \cite{sd}, have enabled more effective modeling of the complex distribution of natural images, and researchers have proposed to leverage the powerful T2I generative priors to achieve more perceptually realistic Real-ISR outcomes. 
Many SD-based methods \cite{stablesr,diffbir,seesr,supir} adopt a ControlNet \cite{controlnet} architecture, extracting features from the LQ image as control signals to guide the generation of HQ image from random noise. Meanwhile, methods have also been proposed to extract semantic information, such as tags \cite{seesr}, short captions \cite{pasd} and long descriptions \cite{supir}, from the LQ image to help the SD model synthesizing more visual details.  
One-step diffusion-based Real-ISR methods \cite{osediff, pisa-sr, adcsr} have also been proposed, which directly take the LQ image as input and finetune the SD model using LoRA \cite{lora} techniques. 

\begin{figure}[t]
  \centering 
  \begin{subfigure}[b]{0.45\textwidth}
    \includegraphics[width=\textwidth]{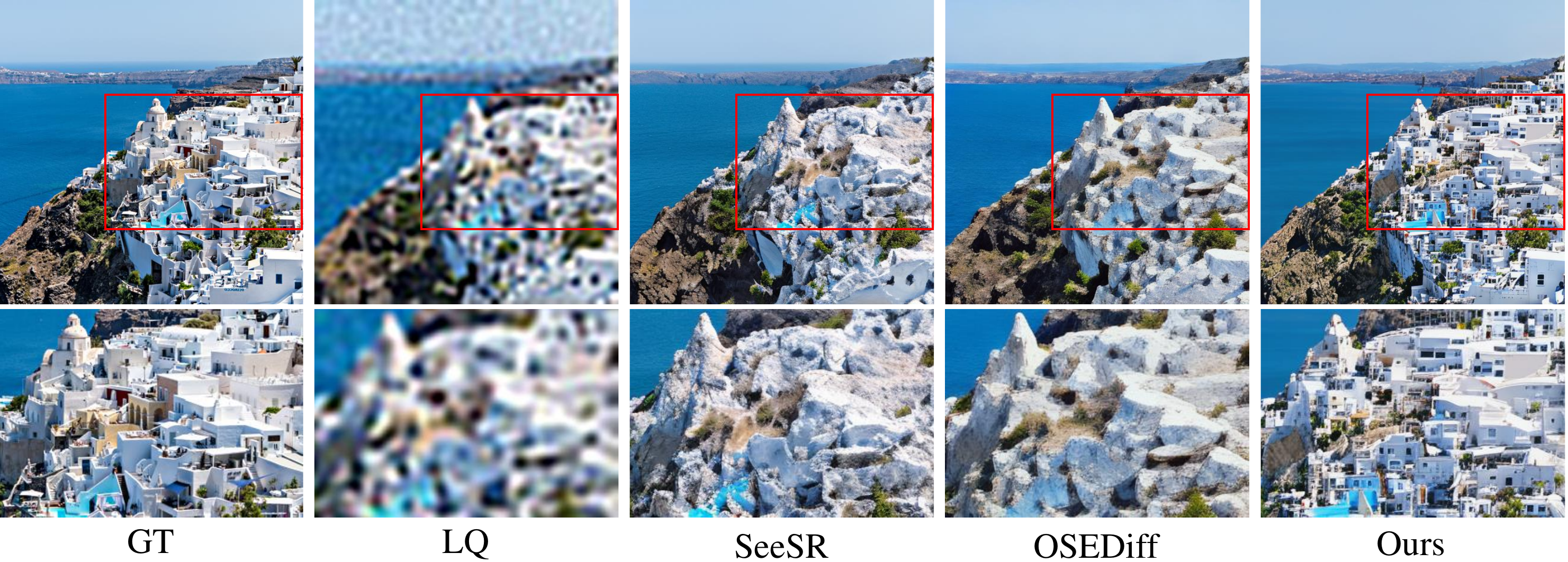}
    \vspace{-5mm}
    \caption{Incorrect content restoration}
    \label{fig:sub1}
  \end{subfigure}
  \begin{subfigure}[b]{0.45\textwidth}
    \includegraphics[width=\textwidth]{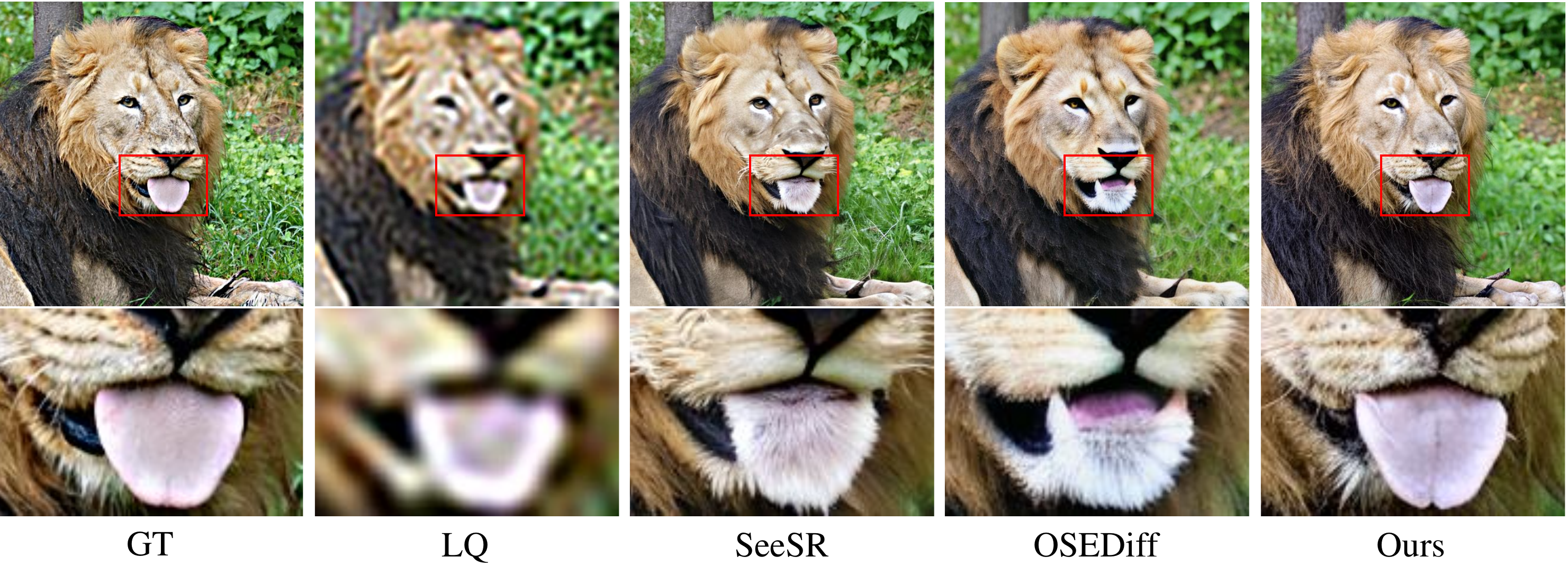}
    \vspace{-5mm}
    \caption{Physically unnatural restoration}
    \label{fig:sub2}
  \end{subfigure}
  \vspace{-1mm}
  \caption{Examples of of (a) inaccurate and (b) unnatural restoration of current SD-based Real-ISR methods, such as SeeSR\cite{seesr} and OSEDiff \cite{osediff}, in complex and heavily degraded scenes.}
  \label{fig:motivation}
  \vspace{-6mm}
\end{figure}

Despite the significant progress achieved in SD-based Real-ISR, the existing models exhibit certain limitations, particularly in handling complex scenes with multiple objects or severe degradation. First, they may lack global understanding of the image content, leading to the generation of semantically inaccurate details. As shown in ~\cref{fig:sub1}, the state-of-the-art models SeeSR \cite{seesr} and OSEDiff \cite{osediff} misidentify buildings as cliffs, leading to incorrect restoration of the scene. Second, existing models lack strong perception of the local structural relationships in natural images, leading to semantically coherent but physically unnatural appearance. As shown in ~\cref{fig:sub2}, SeeSR and OSEDiff erroneously synthesize the lion's tongue with fur. 
These limitations highlight the need for a new Real-ISR model, which can simultaneously perceive, understand and restore the image with complex contents.

The recent advances in autoregressive multimodal large language models (MLLMs) \cite{llava, Chameleon, lumina-mgpt, showo, transfusion} may offer a promising solution to the above-mentioned problem. Leveraging the large-capacity network trained on extensive multimodal datasets, these models excel at modeling the joint distribution of texts and images, integrating image understanding and generation within a unified framework, while the autoregressive generation process can model the image local structure relationships, providing a strong prior for local structure perception.
Based on these facts, we present a framework, namely \textbf{PURE}, to \textbf{P}erceive, \textbf{U}nderstand, and \textbf{RE}store the HQ image from its LQ observation by fine-tuning a pre-trained MLLM. 

Specifically, by taking the LQ image as input, we implement instruction tuning on the Lumina-mGPT \cite{lumina-mgpt} to predict image degradation characteristics, semantic descriptions, and consequently the HQ output within a single model. On the one hand, our model can effectively utilize the  perceived image degradation and semantic understanding features during the image generation process. On the other hand, the autoregressive generation process can explicitly consider the relationships between previously generated image tokens and the next tokens to be generated, alleviating the issue of physically unnatural appearance. Furthermore, we propose a entropy-based Top-$k$ sampling strategy, which adaptively chooses the optimal Top-$k$ value based on the entropy during inference, further optimizing the local structure preservation.

Our contributions are summarized as follows. First, we propose, for the first time to our best knowledge, an autoregressive multimodal learning framework, which integrates global understanding and local perception, for robust Real-ISR. Second, we develop an entropy-based Top-$k$ sampling strategy during inference to better preserve image structures while generating realistic details. Finally, our experimental results demonstrate the potential of PURE for Real-ISR tasks, particularly in complex  scenarios.

\section{Related Work}
\label{sec:related}
\textbf{Real-World Image Super-Resolution}.
Traditional Real-ISR methods \cite{srcnn, edsr, rcan, swinir, hat} typically adopt discriminative networks trained with pixel-wise $L_1$ or $L_2$ losses to recover high-frequency details, but often produce over-smoothed results. To improve perceptual quality, GAN-based approaches~\cite{srgan, realesrgan} leverage adversarial losses to synthesize realistic textures, albeit at the cost of unstable training and artifacts. Recent advances focus on exploiting pre-trained text-to-image (T2I) diffusion models, such as SD~\cite{sd}, as generative priors. Some methods employ a  ControlNet-based architecture~\cite{controlnet} to guide SD generation using LQ image features as conditional signals, while some other methods enhance semantic alignment by extracting text descriptions~\cite{supir, pasd} or tags~\cite{seesr} from LQ inputs. One-step diffusion variants~\cite{osediff, pisa-sr, adcsr} further accelerate inference by directly mapping LQ to HQ images through LoRA-finetuned SD-UNet. Despite improved perceptual quality, these methods struggle with semantic inaccuracies and unnatural local structures in complex scenes due to limited global understanding and local relationship modeling.

\noindent \textbf{Multimodal Large Language Models}.
Recent advancements in MLLMs \cite{llava, Chameleon, lumina-mgpt, showo, transfusion} have expanded the capabilities of traditional language models to process and generate cross-modal content. Early works, such as LLaVA \cite{llava} and BLIP \cite{blip}, focused on enhancing multimodal understanding by aligning pretrained vision encoders (\eg, CLIP \cite{clip}) with LLMs like LLaMA \cite{llama}, enabling tasks such as visual question answering and image captioning. However, these models primarily emphasized perception, lacking native generative abilities for visual content. To address multimodal generation, approaches like Show-o \cite{showo} and Transfusion \cite{transfusion} integrate autoregressive text modeling with diffusion-based image decoders (\eg, SD), where the LLM predicts text tokens and relies on external diffusion models to synthesize images. Although effective, this hybrid paradigm introduces dependency on separate generative frameworks, leading to inconsistent outputs and limited end-to-end learning. In contrast, unified understanding and generation is pioneered by multimodal models such as Chameleon \cite{Chameleon}, Lumina-mGPT \cite{lumina-mgpt} and Emu3 \cite{emu3}, which treat images as discrete tokens within an autoregressive framework, enabling seamless multimodal sequence modeling for text and images. In this work, we employ Lumina-mGPT as the MLLM for Real-ISR tasks. 
\section{Method}
\label{sec:method}

\subsection{Preliminaries}
\label{sec:preliminaries}
Chameleon \cite{Chameleon} is a representative MLLM, which uses next-token prediction in a discrete space to handle texts and images. Lumina-mGPT \cite{lumina-mgpt} further enhances Chameleon by enabling photorealistic text-to-image generation through supervised fine-tuning. In our work, we leverage Lumina-mGPT as the base model for Real-ISR, leveraging its strong image synthesis and understanding priors.

\noindent \textbf{Vision Tokenizer}. 
Lumina-MGPT \cite{lumina-mgpt} unifies vision-language understanding and generation through a VQGAN-based \cite{vqgan} tokenizer, which converts continuous images into symbolic sequences. It integrates three functional components: an encoder compressing visual inputs with downsampling factor $f$, a quantizer discretizing latent vectors by nearest-codebook matching with size $N$, and a decoder reconstructing images from token-codebook mappings. In Real-ISR, the tokenizer will directly impact reconstruction quality: lower $f$ preserves spatial details by reducing compression, while larger $N$ enhances feature expressiveness through fine-grained codebook representations.

\begin{figure*}[t]
    \centering
    \includegraphics[width=\textwidth]{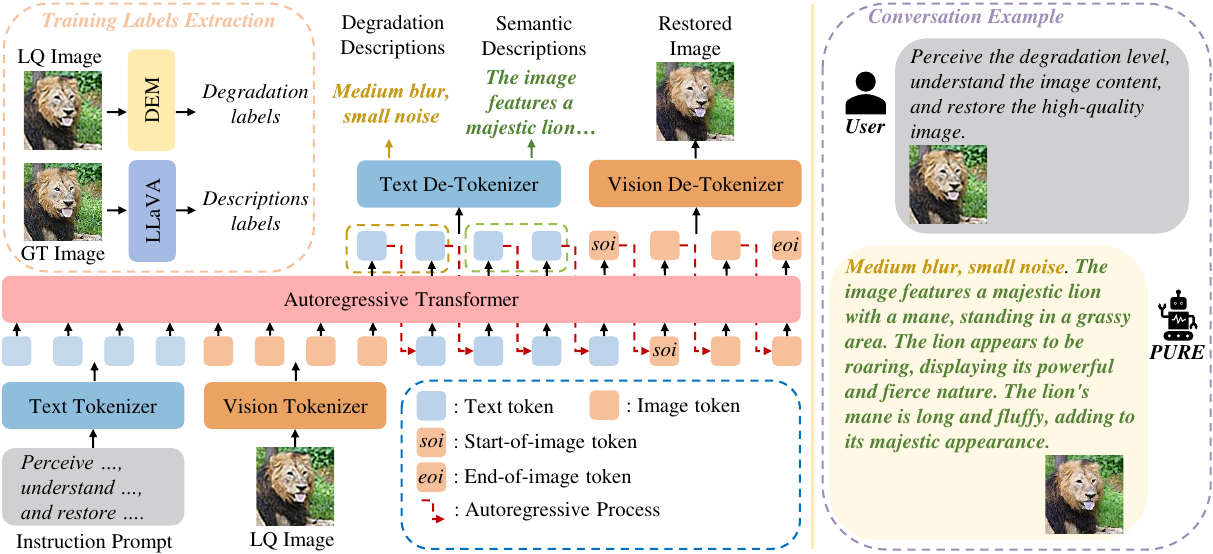}
    \caption{Overview of PURE. We utilize VQGAN's encoder as the vision tokenizer to generate discrete tokens of the given LQ image. The instruction is passed to the text tokenizer to generate text tokens, which are then concatenated with image tokens. Subsequently, the backbone transformer performs autoregressive prediction of the image degradation, semantic understanding and HQ image tokens. Finally, the output image tokens are passed to the vision de-tokenizer and decoded to the restored image.}
    \label{fig:pipeline}
    \vspace{-3mm}
\end{figure*}

\noindent \textbf{Training}. Multimodal large language models process concatenated text-image token sequences through autoregressive learning. The objective maximizes the likelihood of target sequence $\mathbf{y} = \{y_1,...,y_T\}$ via conditional probability factorization $P(\mathbf{y}|\mathbf{x}) = \prod_{t=1}^T P(y_t|\mathbf{c}_{<t},\mathbf{x})$, where $\mathbf{c}_{<t} = \{y_1,...,y_{t-1}\}$ denotes preceding context, and $\mathbf{x}$ represents conditional inputs (\eg, low-quality images). Optimization minimizes stepwise CrossEntropy between predictions and targets: 
$\mathcal{L} = CrossEntropy(P(y_t|\mathbf{c}_{<t},\mathbf{x}), y_t)$.

\noindent \textbf{Inference}. Autoregressive generation samples tokens $y_t \sim P(y_t|\mathbf{c}_{<t},\mathbf{x})$ sequentially. To balance diversity and quality, Top-$k$ sampling truncates the $k$-highest candidates before renormalization $y_t \sim P_{\text{Top-}k}(y_t|\mathbf{c}_{<t},\mathbf{x})$, where hyperparameter $k \in \mathbb{Z}^+$ governs the candidate pool size.

\subsection{Architecture of PURE} 

To address the issue that SD-based Real-ISR models tend to generate inaccurate and unnatural reconstructions in complex and/or heavily degraded scenes, we propose PURE, a robust Real-ISR framework based on autoregressive multimodal generative models. 
As illustrated in~\cref{fig:pipeline}, our model architecture comprises three core components: a text tokenizer, a vision tokenizer, and a decoder-only Transformer backbone. The Transformer we employed primarily follows the design principles of Lumina-mGPT, which consists of RMSNorm \cite{rmsnorm} and QK-Norm \cite{Chameleon} for normalization, SwiGLU \cite{swiglu} as the activation function, and rotary positional embeddings (RoPE) \cite{rope} for positional encoding.

As discussed in \cref{sec:preliminaries}, the vision tokenizer plays a critical role in extracting discrete latent representations of visual inputs, while its quality directly impacts the performance of downstream vision tasks. Although Lumina-mGPT utilizes a VQGAN with $f$ of 16 and $N$ of 8,192, allowing it to encode a 512$\times$512 image into a sequence of 1,024 discrete tokens, the high compression ratio and limited codebook capacity lead to substantial information loss during the extraction of discrete image representations. This loss is particularly detrimental to image restoration tasks, where preserving fine-grained image structure and details is critical. To mitigate this issue, we propose to adopt a more advanced vision tokenizer. Specifically, we employ the VQGAN introduced in LlamaGen \cite{llamagen}, which features a reduced $f$ of 8 and an expanded $N$ of 16,384. This configuration can significantly reduces information loss during encoding, thereby enhancing the quality of reconstructed images in our Real-ISR task.

For the text tokenizer, we retain the Byte Pair Encoding (BPE) tokenizer \cite{Chameleon} from Lumina-mGPT with a vocabulary size of 65,536. Since the vocabulary includes image codebook tokens, expanding VQGAN's codebook size needs a corresponding expansion in BPE tokenizer's vocabulary. We therefore expand the vocabulary size from 65,536 to 73,728 to accommodate the additional discrete image tokens. To accommodate the expanded VQGAN codebook and BPE tokenizer vocabulary, we resize the model's embedding layer (which maps discrete symbols to continuous vector representations), the output layer, and a classification head that predicts distributions over the vocabulary. 

\subsection{Training of PURE }

As illustrated in~\cref{fig:pipeline}, PURE takes the instruction and an LQ image as inputs and sequentially outputs degradation levels, semantic descriptions, and HQ results.
Specifically, given an LQ image, the pipeline is guided by the following instruction: “\textit{Perceive the degradation level, understand the image content, and restore the high-quality image.}” This instruction is first processed by a text tokenizer, converting it into discrete text tokens. Simultaneously, the LQ image is transformed into discrete image tokens via a vision tokenizer. Both sets of tokens are fed into the network as inputs, enabling the model to execute the sequential tasks of perception, understanding, and restoration.

\noindent \textbf{Perception}. 
The perception module initiates the pipeline by quantitatively analyzing the degradation of the input LQ image. We employ a Degradation Estimation module (DEM) in MM-RealSR+ \cite{dem}, which evaluates the degradation of the LQ image and assigns scores for both noise and blur. These scores are categorized into three discrete levels: small, medium, and large, which are then explicitly encoded as \textit{degradation labels} during training. This phase provides quantitative degradation priors for subsequent phases.

\noindent \textbf{Understanding}. 
The phase of understanding constitutes a semantic-aware component in our PURE framework to extract and formalize contextual scene semantics from image content. To obtain these descriptions, we first utilize LLaVa \cite{llava} to generate structured textual descriptions of the ground-truth (GT) images of the LQ inputs. These semantic descriptions are then used to supervise the training of our PURE model by establishing text-visual correlations. 

\noindent \textbf{Restoration}. The restoration phase synthesizes HQ images via autoregressive token generation, which is guided by 3 key elements: the degradation level, the semantic context, and the structural coherence of the evolving sequence. Integrating these elements, we ensure that the reconstructed image can not only address the degradation but also align with the semantics of the scene, resulting in content coherent and visually appealing Real-ISR outputs.

\noindent \textbf{Training Loss}. During training, we utilize the next-token prediction objective, optimizing a standard cross-entropy loss to jointly model the distribution of text and image tokens. The training loss is: 
\begin{equation}
\mathcal{L} = CrossEntropy\left(y_t, p_\theta(x_t \mid x_{<t}, I, X_{\text{LQ}})\right),
\end{equation}
where $p_\theta(x_t \mid x_{<t}, I, X_{\text{LQ}})$ represents the probability distribution of the model for generating the \( i \)-th token, conditioned on the input instruction $I$, LQ image tokens $X_{\text{LQ}}$, and all previously generated tokens $x_{<t}$.

\subsection{Inference with Entropy-based Sampling}
\label{sec:dts}

As mentioned in ~\cref{sec:preliminaries}, we adapt the Top-$k$ sampling strategy during inference. Specifically, a smaller $k$ value restricts the sampling space and can preserve input image structures. In contrast, increasing $k$ can result in images with richer details. Conventional approaches typically employ a global Top-$k$ setting for all image tokens. However, in image restoration tasks, different regions exhibit varying requirements for detail richness. For example, areas with strong edges require fewer details, and a smaller $k$ value is appropriate. Conversely, regions with fine textures (\eg, grasses, animal hairs) benefit from richer details, necessitating a larger $k$ value. The challenge is then turned to how to determine the structural uncertainty of different image regions so that we can adaptively configure a tailored Top-$k$ value for each image token.

\begin{figure}[t]
    \centering
    \includegraphics[width=0.48\textwidth]{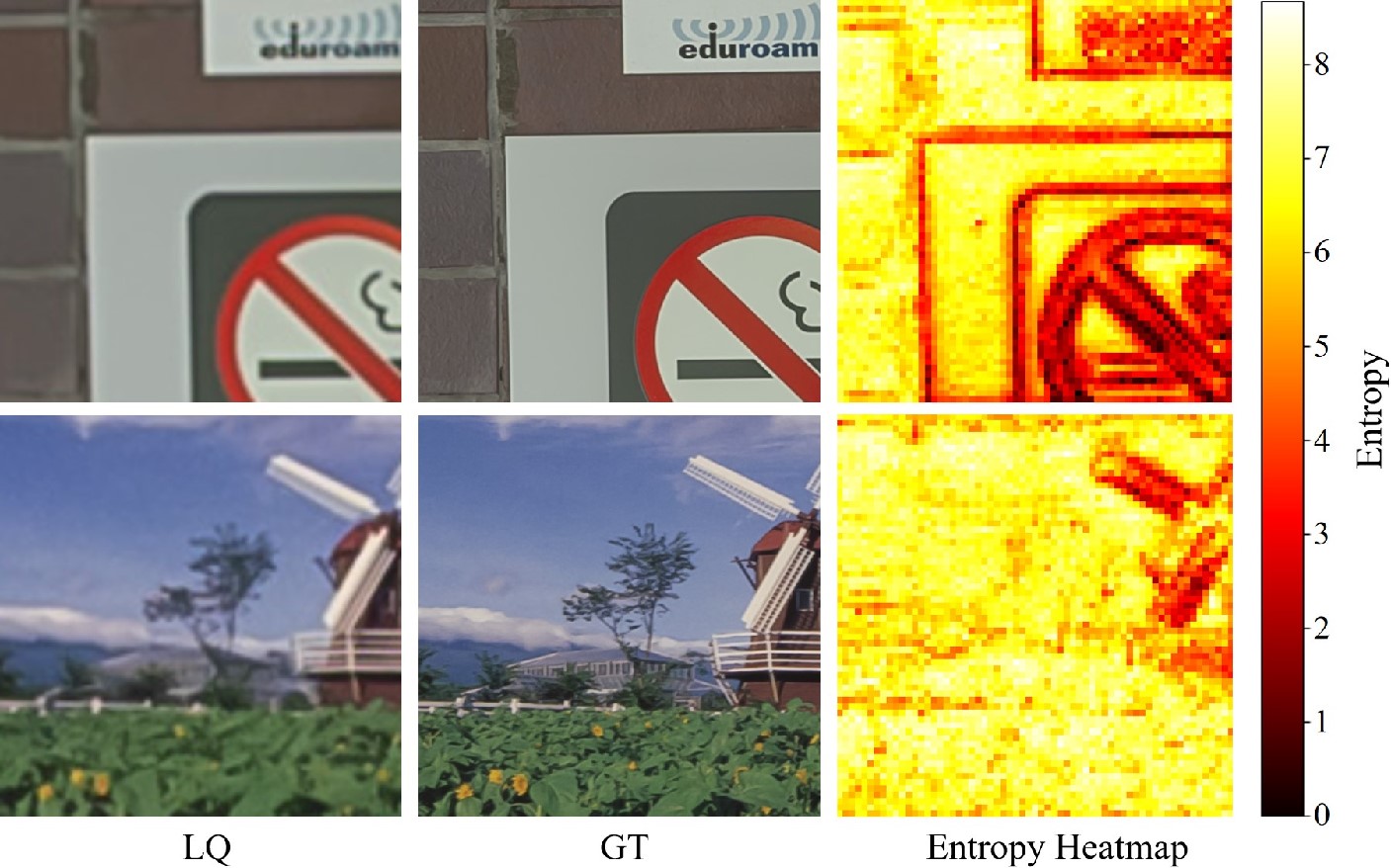}
    \vspace{-7mm}
    \caption{Visualization of entropy distribution in image token generation. The entropy heatmap shows token-wise uncertainty, where darker regions indicate higher entropy values and lighter regions represent lower entropy values.}
    \label{fig:entropy}
    \vspace{-5mm}
\end{figure}

In this work, we use entropy, a concept that can quantify uncertainty in probability distributions, as the uncertainty measure, which is defined as $H_t = -\sum_{i=1}^{n} p_i \log p_i$. By exploring the entropy of each token during generation, we observe a strong correlation between entropy and structural patterns of the generated image. As shown in~\cref{fig:entropy}, high-entropy tokens typically correspond to regions with complex structures or varied textures, while low-entropy tokens are often associated with areas with sharp edges. Based on this insight, we dynamically adjust the Top-$k$ value based on the entropy of the current image token. The dynamic Top-$k$ value \( k_t \) at step \( t \) is calculated as:
\begin{equation}
k_t = k_{\text{min}} + (k_{\text{max}} - k_{\text{min}}) \times \sigma\left({H_t - \mu}\right),
\end{equation}
where \( \sigma(\cdot) \) is the sigmoid function, \( \mu \) is the standard deviation of entropy observed during inference, and \( k_{\text{min}} \) and \( k_{\text{max}} \) are the predefined minimum and maximum values for Top-$k$ sampling during image token generation. 
Our proposed entropy-based Top-$k$ sampling strategy can balance fidelity and diversity in image token generation, significantly improving the perceptual quality of restored images, particularly in complex scenes with diverse spatial structures.
\section{Experiments}
\label{sec:exp}

\subsection{Experimental Setup}
\noindent \textbf{Training  Datasets}. Following  \cite{seesr, osediff}, we train PURE on LSDIR \cite{lsdir} and a subset of 10,000 high-resolution facial images from FFHQ \cite{ffhq}. To generate realistic LQ-HQ image pairs for training, we apply the degradation pipeline from Real-ESRGAN \cite{realesrgan}, which introduces blur kernels, noise perturbations, and JPEG compression artifacts. 

\noindent \textbf{Test Datasets}. For comprehensive evaluation of PURE, we employ one synthetic dataset and three real-world datasets for testing. (1) First, we resize the shortest side of the OST validation set \cite{ost} to 512 pixels and perform center-cropping to obtain 512$\times$512 resolution images as the GTs. Using the same degradation pipeline as in SeeSR \cite{seesr}, we generate LQ-HQ image pairs, collectively referred to as OST-Val, which serves as our synthetic test set. (2) Second, we employ the real-world datasets RealSR \cite{realsr} and RealLQ250 \cite{dreamclear} for testing. 
(3) On the above three datasets, we perform $4\times$ Real-ISR experiments as in most of the existing methods \cite{seesr,pasd,osediff}, where the LQ image is of 128$\times$128 resolution. In this paper, we further evaluate PURE for 2$\times$ Real-ISR using the DrealSR \cite{drealsr} dataset to achieve a wider field of view that better aligns with real-world multi-object scenarios. We center-crop the GT images from the DrealSR dataset to 512$\times$512 while correspondingly center-cropping the LQ images to 256$\times$256. 

\noindent \textbf{Compared Methods}. We compare PURE against state-of-the-art diffusion-based Real-ISR methods, including StableSR \cite{stablesr}, PASD \cite{pasd}, DiffBIR \cite{diffbir}, SeeSR \cite{seesr} and OSEDiff \cite{osediff}. All competing methods utilize comparable size training data, ensuring a fair comparison.

\noindent \textbf{Evaluation Metrics}. To holistically assess restoration quality, we employ a combination of reference-based and no-reference metrics. Pixel-level fidelity is measured through PSNR and SSIM \cite{ssim} computed on the Y channel in YCbCr color space. Perceptual similarity is evaluated using LPIPS \cite{lpips} and DISTS \cite{dists}, which compare deep feature representations between restored and GT images. FID \cite{fid} quantifies the statistical distribution alignment between restored and reference images. For blind quality assessment without reference images, we incorporate NIQE \cite{niqe}, MANIQA \cite{maniqa}, MUSIQ \cite{musiq}, CLIPIQA \cite{clipiqa} and TOPIQ \cite{topiq} to evaluate image naturalness and structural integrity.

\noindent \textbf{Implementation Details}. We train PURE using the AdamW optimizer with an initial learning rate of \(4 \times 10^{-4}\), following a cosine decay schedule during training. The model is trained on the entire dataset for 2 epochs with a batch size of 128, including a warmup phase spanning the first 0.5 epochs to stabilize early-stage optimization. Training is conducted on 512$\times$512 resolution images with 32 NVIDIA A800 GPUs. During the inference process, the dynamic Top-$k$ sampling parameters for image token generation are configured as \(k_{\text{min}}\!=\!1\), \(k_{\text{max}}\!=\!2000\) with \(\mu\!=\!5\), while a conservative Top-\(k\!=\!1\) strategy is implemented for text token generation to ensure optimal textual accuracy. We maintained the inference parameters used in Lumina-mGPT, setting CFG \cite{lumina-mgpt} to 0.8 and temperature to 0.9.


\subsection{Comparison with State-of-the-Arts}

\begin{table}[t]
\centering
\caption{Comparative analysis of PSNR (dB) between the VAE used in SD-based methods and the VQGAN used in our PURE, with performance differences highlighted in red.}
\vspace{-3mm}
\label{tab-tokenizer}
\small
\resizebox{\linewidth}{!}{
\begin{tabular}{c|lll}
\toprule
Tokenizers & OST-Val & RealSR & DrealSR \\ 
\midrule
VAE        & 25.00 & 24.43  & 36.20 \\
\addlinespace[2pt]
VQGAN      & 24.29 (\textcolor{red}{-0.71}) & 23.53 (\textcolor{red}{-0.90})  & 33.59 (\textcolor{red}{-2.62}) \\
\bottomrule
\end{tabular}
\vspace{-7mm}
}
\end{table}

\noindent \textbf{Quantitative Comparisons}. 

\begin{table*}[t]
\centering
\caption{Quantitative comparison between PURE and state-of-the-art diffusion-based methods. The best and second best results of each metric are highlighted in \textcolor{red}{\textbf{red}} and \textcolor{blue}{\textbf{blue}}, respectively.}
\vspace{-2mm}
\resizebox{\linewidth}{!}{
\begin{tabular}{@{}c|c|cccccccccc@{}}
\toprule
Datasets                   & Methods             & PSNR↑ & SSIM↑  & LPIPS↓ & DISTS↓  & FID↓   & NIQE↓  & MUSIQ↑ & MANIQA↑ & CLIPIQA↑ & TOPIQ↑\\ \midrule
\multirow{6}{*}{OST-Val}   & StableSR      & 19.97 & 0.4638 & 0.3367 & 0.2007 &\textbf{\textcolor{red}{51.11}} & \textbf{\textcolor{red}{3.7669}} & \textbf{\textcolor{blue}{74.38}} & 0.6009 & 0.7194 & 0.7789  \\
                           & DiffBIR       & 20.47 & 0.4694 & 0.3622 & 0.2059 & 61.28 & 4.5477 & 73.98 & \textbf{\textcolor{blue}{0.6377}} & \textbf{\textcolor{blue}{0.7378}} & \textbf{\textcolor{blue}{0.7983}}  \\
                           & SeeSR         & \textbf{\textcolor{blue}{20.71}} & \textbf{\textcolor{blue}{0.5004}} & \textbf{\textcolor{blue}{0.3274}} & \textbf{\textcolor{blue}{0.1935}} & 56.32 & 3.9399 & 74.03 & 0.5920 & 0.6900 & 0.7945  \\
                           & PASD          & \textbf{\textcolor{red}{20.95}} & \textbf{\textcolor{red}{0.5053}} & 0.4121 & 0.2190 & 66.92 & 4.8305 & 69.77 & 0.4289 & 0.5141 &  0.6307 \\
                           & OSEDiff       & 20.59 & 0.4954 & \textbf{\textcolor{red}{0.3133}} & \textbf{\textcolor{red}{0.1926}} & \textbf{\textcolor{blue}{52.11}} & \textbf{\textcolor{blue}{3.8947}} & 74.30 & 0.6102 & 0.6709 & 0.7755  \\
                           & PURE          & 18.52 & 0.4181 & 0.3863 & 0.2052 & 55.19 & 4.7977 & \textbf{\textcolor{red}{74.81}} & \textbf{\textcolor{red}{0.6424}} & \textbf{\textcolor{red}{0.7581}} & \textbf{\textcolor{red}{0.8148}}  \\
                           \midrule
\multirow{6}{*}{RealSR}    & StableSR & 24.64 & 0.7079 & 0.3004 & 0.2141 & 128.66 & 5.8823 & 72.19 & 0.4054 & 0.6234 & 0.6809  \\
                           & DiffBIR        & 24.82 & 0.6495 & 0.3656 & 0.2400 & 131.01 & 5.8477 & 72.50 & \textbf{\textcolor{red}{0.4729}} & \textbf{\textcolor{red}{0.7059}} &  \textbf{\textcolor{blue}{0.7313}} \\
                           & SeeSR         & 25.14 & 0.7210 & 0.3008 & 0.2224 & 125.61 & \textbf{\textcolor{red}{5.3913}} & \textbf{\textcolor{blue}{72.61}} & 0.4203 & 0.6705 & \textbf{\textcolor{red}{0.7354}}  \\
                           & PASD          & \textbf{\textcolor{red}{26.63}} & \textbf{\textcolor{red}{0.7660}} & \textbf{\textcolor{red}{0.2870}} & \textbf{\textcolor{red}{0.2034}} & \textbf{\textcolor{blue}{125.26}} & 5.9892 & 70.01 & 0.3382 & 0.4869  & 0.5988 \\
                           & OSEDiff       & \textbf{\textcolor{blue}{25.15}} & \textbf{\textcolor{blue}{0.7341}} & \textbf{\textcolor{blue}{0.2920}} & 0.2127 & \textbf{\textcolor{red}{123.73}} & 5.6442 & \textbf{\textcolor{red}{72.95}} & 0.4507 & 0.6686 & 0.7106  \\
                           & PURE          &  22.83 & 0.6079 & 0.3821 & 0.2458 & 188.33 & \textbf{\textcolor{blue}{5.4090}} & 72.37 & \textbf{\textcolor{blue}{0.4527}} & \textbf{\textcolor{blue}{0.6817}}  & 0.7165 \\
                           \midrule
\multirow{6}{*}{DrealSR}   & StableSR       & 25.86 & 0.7380 & 0.2875 & 0.1951 & 95.71 & 5.9033 & \textbf{\textcolor{blue}{73.55}} & 0.4399 & 0.7021 &  0.7408 \\
                           & DiffBIR         & 26.82 & 0.7035 & 0.3389 & 0.2179 & 80.11 & 5.6977 & 73.23 & \textbf{\textcolor{blue}{0.5007}} & \textbf{\textcolor{blue}{0.7225}}  & \textbf{\textcolor{blue}{0.7608}}  \\
                           & SeeSR           & \textbf{\textcolor{blue}{27.71}} & 0.7676 & 0.2768 & 0.2033 & 93.80 & \textbf{\textcolor{red}{5.4440}} & 73.45 & 0.4808 & 0.7046  &  \textbf{\textcolor{red}{0.7673}} \\
                           & PASD            & \textbf{\textcolor{red}{29.73}} & \textbf{\textcolor{red}{0.8326}} & \textbf{\textcolor{red}{0.2099}} & \textbf{\textcolor{red}{0.1643}} & \textbf{\textcolor{red}{67.16}} & 5.6899 & 72.40& 0.4204 & 0.6268  & 0.6979 \\
                           & OSEDiff          & 26.60 & \textbf{\textcolor{blue}{0.7693}} & \textbf{\textcolor{blue}{0.2633}} & \textbf{\textcolor{blue}{0.1882}} & 94.59 & 5.7724 & \textbf{\textcolor{red}{73.98}} & 0.4978 & 0.7114 & 0.7345 \\
                           & PURE          & 24.73 & 0.6459 & 0.3452 & 0.2076 & \textbf{\textcolor{blue}{79.41}} & \textbf{\textcolor{blue}{5.4846}} & 73.28 & \textbf{\textcolor{red}{0.5251}} & \textbf{\textcolor{red}{0.7259}} & 0.7560  \\
                           \midrule
                           \bottomrule
\end{tabular}
}
\label{tab:main}
\vspace{-3mm}
\end{table*}

In all of our competing methods, the variational auto-encoder (VAE) \cite{vae} is used to convert the input image into latent space for diffusion, while in our PURE, a VQGAN is used to convert images into discrete tokens for processing. Both VAE and VQGAN will lose image details, but VAE is more advantageous for its continuous representation. \cref{tab-tokenizer} quantify the fidelity (in terms of as PSNR) of GT images on the three test sets, which is actually the PSNR upper bounds of Real-ISR performance. Taking DrealSR as an example, VAE outperforms VQGAN by 2.61 dB. Taking this fact into account, we investigate whether PURE, which relies on VQGAN for autoregressive token generation, can overcome this deficiency and show competitive Real-ISR performance.   

\begin{table}[t]
\centering
\caption{Quantitative results on the RealLQ250 dataset. The best and second best results of each metric are highlighted in \textcolor{red}{\textbf{red}} and \textcolor{blue}{\textbf{blue}}, respectively.}
\vspace{-2mm}
\resizebox{\linewidth}{!}{
\begin{tabular}{@{}c|ccccc@{}}
\toprule
Methods   & NIQE↓  & MUSIQ↑ & MANIQA↑ & CLIPIQA↑ & TOPIQ↑ \\ \midrule
StableSR  & 5.9122 & 73.75  & 0.5075  & 0.7273   & 0.7431 \\
DiffBIR   & 4.7547 & 73.68  & \textbf{\textcolor{red}{0.5971}} & \textbf{\textcolor{blue}{0.7669}} & \textbf{\textcolor{blue}{0.7629}} \\
SeeSR     & \textbf{\textcolor{blue}{4.2094}} & 73.91  & 0.5241  & 0.7237   & 0.7621 \\
PASD      & 4.3238 & 72.14  & 0.4388  & 0.6200   & 0.6578 \\
OSEDiff   & \textbf{\textcolor{red}{4.1491}} & \textbf{\textcolor{red}{74.21}} & 0.5263  & 0.7101   & 0.7285 \\
PURE      & 4.5054 & \textbf{\textcolor{blue}{74.10}} & \textbf{\textcolor{blue}{0.5409}} & \textbf{\textcolor{red}{0.7711}} & \textbf{\textcolor{red}{0.7751}} \\
\bottomrule
\end{tabular}
}
\vspace{-5mm}
\label{tab:reallq250}
\end{table}

~\cref{tab:main,tab:reallq250} presents a comprehensive quantitative comparison between PURE and state-of-the-art diffusion-based approaches on the four test datasets. (Note that RealLQ250 does not have GTs so that we can only compute the no-reference metrics.) Although PURE, as expected, show inferior results in full-reference based metrics such as PSNR, SSIM, \etc, it demonstrates highly competitive performance across no-reference metrics. This substantiates that PURE, via effective perception and understanding of the input LQ image, can exhibit robust image restoration capability and produce HQ images aligning with human visual preferences. We argue that this achievement is non-trivial considering the significant loss of details in the discrete tokenization process. On the other hand, such results imply the significant potentials of autoregressive multimodal generative models in real-world image restoration tasks.

\begin{figure*}[t]
  \centering
  \includegraphics[width=1.0\linewidth]{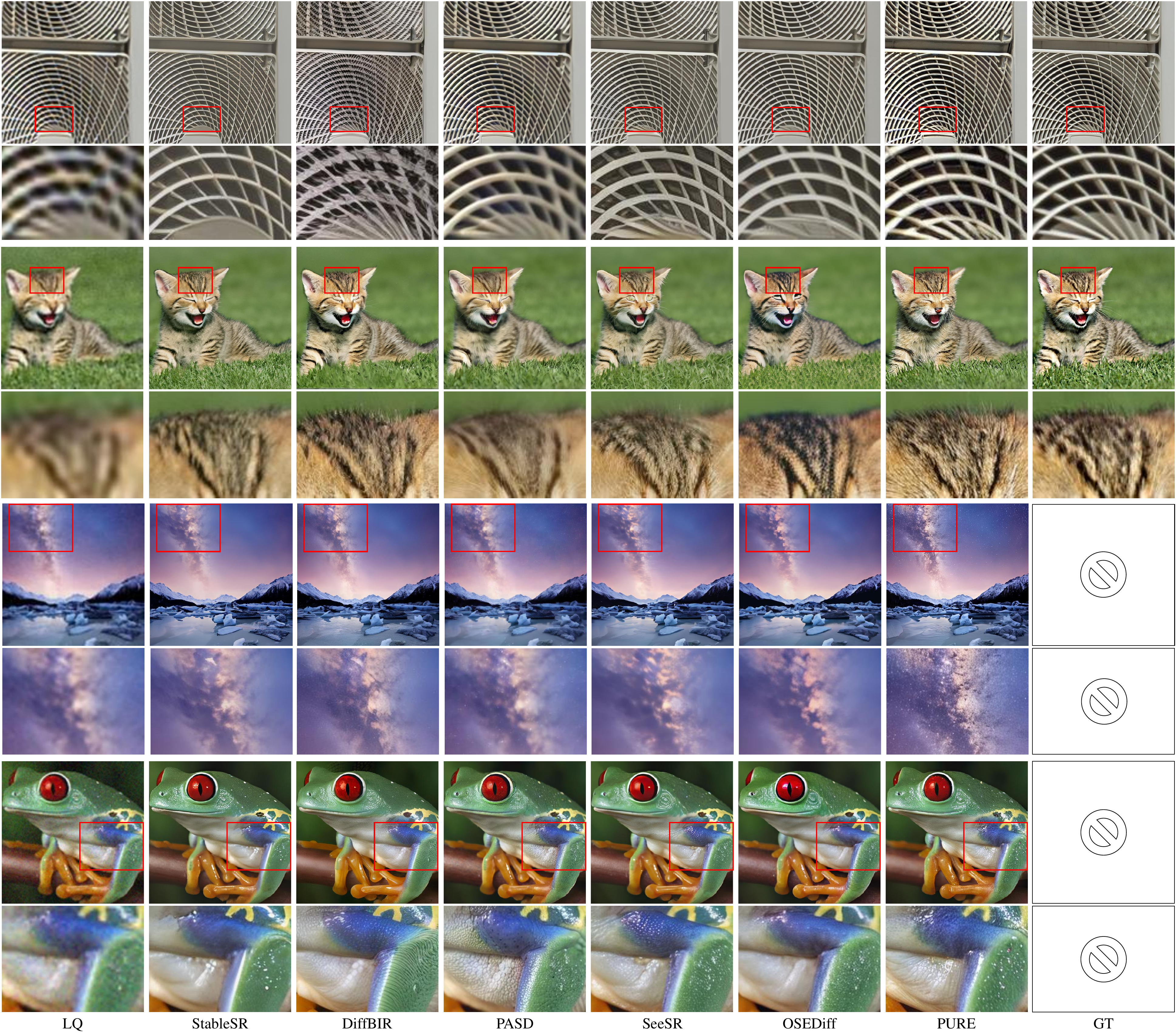}
\vspace{-5mm}
  \caption{Qualitative comparisons of different Real-ISR methods. Please zoom in for a better view.}
  \label{fig:visual}
  \vspace{-3mm}
\end{figure*}

\noindent \textbf{Qualitative Comparison}. 
We present qualitative comparisons with other Real-ISR methods in ~\cref{fig:visual}. As shown in the first case, our PURE approach successfully recovers the most authentic and accurate details, while the other methods yield either over-smoothed or incorrect textures. Similar observation can be found in the second case, where our approach generates more natural results with significantly reduced artifacts compared to the competing methods. For the remaining cases, PURE also shows advantages in terms of visual quality, producing more realistic texture details while maintaining better overall image fidelity. More visual results are provided in ~\cref{sec:supp1}.

\begin{figure}[htbp]
  \centering
  \includegraphics[width=0.8\linewidth]{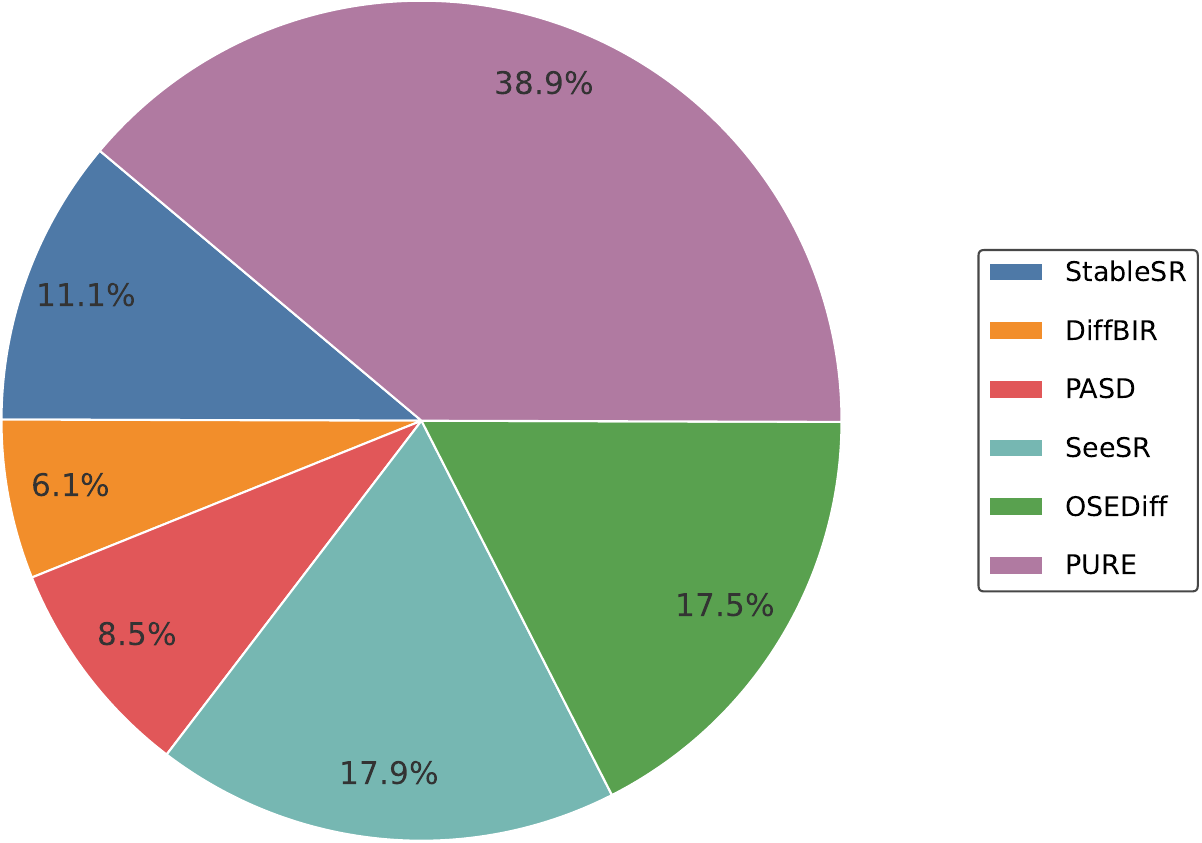}
  \vspace{-3mm}
  \caption{User study results on PURE and five state-of-the-art diffusion-based Real-ISR methods (StableSR, PASD, DiffBIR, SeeSR, OSEDiff) with 50 real-world images.}
  \label{fig:userstudy}
  \vspace{-5mm}
\end{figure}

\noindent \textbf{User Study}. To further evaluate the performance of competing methods, we conduct a user study. 
We randomly select 50 real-world LQ images from RealSR, DrealSR and RealLQ250, then employ PURE and five state-of-the-art diffusion-based Real-ISR approaches (StableSR, PASD, DiffBIR, SeeSR and OSEDiff) to enhance them. 
These Real-ISR results, together with the LQ images, are presented to volunteers, who are asked to select the best restoration output based on two key criteria: 1) content consistency with the LQ input (including color fidelity, structural preservation, and texture accuracy); and 2) visual perceptual quality (encompassing natural appearance and richness of plausible details).
Fifteen volunteers are invited to participate in the study, each making 50 pairwise comparisons over different methods. As demonstrated in ~\cref{fig:userstudy}, our PURE method achieves significantly better preference than diffusion-based methods, with a selection rate of 38.9\%. This result demonstrates the great potential for autoregressive multi-modal models in solving real-world image restoration tasks.  

\noindent \textbf{Complexity Analysis}. We then make a complexity comparison between PURE and the competing methods. As shown in ~\cref{tab:speed}, it is not a surprise that PURE, as an auto-regressive multimodal generative model-based method, has much more parameters and significantly longer inference time than those SD-based methods. 
Specifically, the large amount of parameters are inherited from the Lumina-mGPT model, while the slower inference speed is determined by the sequential processing manner of autoregressive models, which predict the next-token one by one. In contrast, the SD-based approaches can perform diffusion denoising in parallel. However, it is worth mentioning our work is the first one, to our best knowledge, which successfully applies the auto-regressive multimodal generative model to the task of Real-ISR and demonstrates its effectiveness. 

\begin{table}[] \scriptsize
\caption{Complexity comparison among different methods. All methods are tested with an input image of size $512\times512$, and the inference time is measured on an A100 GPU.}
\vspace{-2mm}
\centering
\resizebox{\linewidth}{!}{
\begin{tabular}{@{}c|cccccc@{}}
\toprule
                    & StableSR & DiffBIR & SeeSR  & PASD   & OSEDiff & PURE \\ \midrule
Inference Time (s)  & 11.50    & 2.72    & 4.30   & 2.80   & 0.35    & 256.87   \\
Total Param (M)     & 1410     & 1717    & 2524   & 1900   & 1775    & 7080   \\

\bottomrule
\end{tabular}
}
\label{tab:speed}
\vspace{-4mm}
\end{table}

\begin{table*}[t] \scriptsize
\caption{Ablation study on perception-understanding guidance mechanisms. The best results of each metric are marked in \textbf{bold}.}
\vspace{-3mm}
\centering
\label{tab:text_ablation}
\resizebox{\linewidth}{!}{
\begin{tabular}{@{}c|cccccccccc@{}}
\toprule
Strategy & PSNR↑ & SSIM↑  & LPIPS↓ & DISTS↓  & FID↓   & NIQE↓  & MUSIQ↑ & MANIQA↑ & CLIPIQA↑ & TOPIQ↑ \\ \midrule
\textit{No Guidance} & 23.43 & 0.5588 & 0.4452 & 0.2409 & 191.21 & 5.9722 & 73.23 & 0.4950 & 0.7202 & \textbf{0.7601} \\
\textit{Perception Guidance} & 24.48 & \textbf{0.6510} & 0.3471 & 0.2106 & 83.20 & 5.5573 & 73.15 & 0.5189 & 0.7166 & 0.7410 \\
\textit{Understanding Guidance} & 24.23 & 0.6282 & 0.3937 & 0.2244 & 141.05 & 5.6898 & 73.03 & 0.4955 & 0.7168 & 0.7493 \\
\textit{Full Guidance} & \textbf{24.73} & 0.6459 & \textbf{0.3452} & \textbf{0.2076} & \textbf{79.41} & \textbf{5.4846} & \textbf{73.28} & \textbf{0.5251} & \textbf{0.7259} & 0.7560 \\ \bottomrule
\end{tabular}
\vspace{-6mm}
}
\end{table*}

\begin{table*}[t] \scriptsize
\caption{Ablation study on sampling strategies. The best results of each metric are marked in \textbf{bold}.}
\vspace{-3mm}
\centering
\label{tab:topk_ablation}
\resizebox{\linewidth}{!}{
\begin{tabular}{@{}c|cccccccccc@{}}
\toprule
Strategy & PSNR↑ & SSIM↑  & LPIPS↓ & DISTS↓  & FID↓   & NIQE↓  & MUSIQ↑ & MANIQA↑ & CLIPIQA↑ & TOPIQ↑ \\ \midrule
\textit{Top-1} & \textbf{27.59} & \textbf{0.7945} & \textbf{0.2507} & 0.2173 & 115.39 & 7.9308 & 71.82 & 0.3779 & 0.5291 & 0.6313 \\
\textit{Top-2000} & 24.56 & 0.6379 & 0.3619 & 0.2107 & 80.54 & 5.6159 & \textbf{73.38} & 0.5180 & 0.7169 & 0.7471 \\
\textit{Dynamic Top-$k$} & 24.73 & 0.6459 & 0.3452 & \textbf{0.2076} & \textbf{79.41} & \textbf{5.4846} & 73.28 & \textbf{0.5251} & \textbf{0.7259} & \textbf{0.7560} \\ \bottomrule
\end{tabular}
}
\vspace{-5mm}
\end{table*}

\subsection{Ablation Study}
We perform ablation studies on our proposed perception-understanding strategy and the entropy-based Top-$k$ sampling strategy. The visual comparisons for the ablation study are presented in ~\cref{sec:supp2}.

\noindent \textbf{Perception-Understanding Guidance}. To validate the effectiveness of our sequential perception-understanding-restoration paradigm, we systematically ablate the contribution of each text-guided component on DrealSR. The \textit{Full Guidance} strategy strictly follows our paradigm, sequentially producing degradation scores, semantic descriptions, and restored images. \textit{Perception Guidance} omits semantic captioning but retains degradation estimation, while \textit{Understanding Guidance} removes degradation analysis to focus solely on semantic understanding. \textit{No Guidance} bypasses all textual intermediates, directly mapping low-quality inputs to restored outputs. The results are shown in ~\cref{tab:text_ablation}, from which we can have the following observations.

First, the \textit{No Guidance} method under-performs on most metrics, particularly on reference-based metrics. In comparison, the \textit{Perception Guidance} method shows significant improvements on reference-based metrics, especially PSNR (with a 1.05 dB increase ), demonstrating the effectiveness of adaptive degradation perception for restoration. Meanwhile, \textit{Understanding Guidance} achieves substantial improvements over \textit{No Guidance} on reference-based metrics and some no-reference metrics (NIQE and MANIQA), proving that semantic understanding can help generate realistic details that better align with the original image's semantics.
Finally, \textit{Full Guidance} attains the best performance across most metrics. This hierarchy confirms that explicit degradation modeling enhances structural fidelity, semantic understanding drives realistic detail generation, and their integration through Full Guidance well balances physical accuracy and perceptual authenticity.

\noindent \textbf{Entropy-based Top-$k$ Sampling}. We evaluate the proposed entropy-based dynamic Top-$k$ sampling strategy against fixed Top-$k$ baselines on DrealSR: 1) \textit{Top-1}: Deterministic sampling ($k=1$); 2) \textit{Top-2000}: Maximal stochasticity ($k=2000$). 
As quantified in ~\cref{tab:topk_ablation}, \textit{Top-1} strategy achieves the best PSNR/SSIM metrics as it consistently selects the token with the highest probability, ensuring strong fidelity. However, it performs poorly on most other metrics due to the lack of diversity in generation. In contrast, the \textit{Top-2000} strategy exhibits the worst PSNR/SSIM scores but outperforms \textit{Top-1} on other metrics, benefiting from its diversity. The proposed Dynamic Top-$k$ strategy strikes an effective balance between these two extremes. While achieving good PSNR and SSIM scores, it achieves the best performance on perceptual metrics (DISTS, FID) and no-reference metrics (NIQE, MUSIQ, MANIQA). This demonstrates that our dynamic Top-$k$ sampling approach successfully navigates the trade-off between fidelity and perceptual quality, leveraging adaptive stochasticity to enhance the realism and naturalness of generated results without compromising much structural accuracy.

\section{Conclusion and Limitation}
\label{sec:conclusion}
We proposed PURE, a robust Real-ISR framework that integrated perception, understanding and restoration through instruction-tuning of the autoregressive multimodal generative model Lumina-mGPT. By jointly perceiving degradation strength, generating semantic descriptions, and restoring high-quality image tokens autoregressively, PURE addressed effectively the limitations of existing diffusion-based methods for handling complex scenes with semantic ambiguity and restoring unnatural local structures. The proposed entropy-based Top-k sampling strategy further improved local structure by adaptively adjusting token generation. Experiments showed that PURE effectively preserved both global semantics and local details in challenging real-world scenarios, showcasing the potential of autoregressive multimodal paradigms for low-level vision tasks.

\vspace{+1mm}
\noindent \textbf{Limitations}. 
Due to the large number of parameters in multimodal generative models and the nature of autoregressive model's next-token prediction paradigm, PURE is large in size and slow in inference speed, limiting its application in many resource-constrained devices. In addition, VQGAN is employed as the vision decoder, whose reconstruction accuracy is lower than that of VAE in diffusion models. In future work, we plan to develop distillation techniques and explore parallel decoding strategies to improve computational efficiency, and investigate more effective vision encoders that can be adopted in autoregressive models.

{
    \small
    \bibliographystyle{ieeenat_fullname}
    \bibliography{main}

\begin{thebibliography}{62}
\providecommand{\natexlab}[1]{#1}
\providecommand{\url}[1]{\texttt{#1}}
\expandafter\ifx\csname urlstyle\endcsname\relax
  \providecommand{\doi}[1]{doi: #1}\else
  \providecommand{\doi}{doi: \begingroup \urlstyle{rm}\Url}\fi

\bibitem[sd()]{sd}
Stability.ai.
\newblock \url{https://stability.ai/stable-diffusion}.

\bibitem[Ai et~al.(2024)Ai, Zhou, Huang, Han, Chen, You, and Yang]{dreamclear}
Yuang Ai, Xiaoqiang Zhou, Huaibo Huang, Xiaotian Han, Zhengyu Chen, Quanzeng You, and Hongxia Yang.
\newblock Dreamclear: High-capacity real-world image restoration with privacy-safe dataset curation.
\newblock In \emph{The Thirty-eighth Annual Conference on Neural Information Processing Systems}, 2024.

\bibitem[Cai et~al.(2019)Cai, Zeng, Yong, Cao, and Zhang]{realsr}
Jianrui Cai, Hui Zeng, Hongwei Yong, Zisheng Cao, and Lei Zhang.
\newblock Toward real-world single image super-resolution: A new benchmark and a new model.
\newblock In \emph{ICCV}, pages 3086--3095, 2019.

\bibitem[Chen et~al.(2024{\natexlab{a}})Chen, Li, Wu, Zhang, Chen, Zhang, and Zhang]{adcsr}
Bin Chen, Gehui Li, Rongyuan Wu, Xindong Zhang, Jie Chen, Jian Zhang, and Lei Zhang.
\newblock Adversarial diffusion compression for real-world image super-resolution, 2024{\natexlab{a}}.

\bibitem[Chen et~al.(2024{\natexlab{b}})Chen, Mo, Hou, Wu, Liao, Sun, Yan, and Lin]{topiq}
Chaofeng Chen, Jiadi Mo, Jingwen Hou, Haoning Wu, Liang Liao, Wenxiu Sun, Qiong Yan, and Weisi Lin.
\newblock Topiq: A top-down approach from semantics to distortions for image quality assessment.
\newblock \emph{IEEE Transactions on Image Processing}, 33:\penalty0 2404--2418, 2024{\natexlab{b}}.

\bibitem[Chen et~al.(2023{\natexlab{a}})Chen, Wang, Zhou, Qiao, and Dong]{hat}
Xiangyu Chen, Xintao Wang, Jiantao Zhou, Yu Qiao, and Chao Dong.
\newblock Activating more pixels in image super-resolution transformer.
\newblock In \emph{Proceedings of the IEEE/CVF Conference on Computer Vision and Pattern Recognition}, pages 22367--22377, 2023{\natexlab{a}}.

\bibitem[Ding et~al.(2020)Ding, Ma, Wang, and Simoncelli]{dists}
Keyan Ding, Kede Ma, Shiqi Wang, and Eero~P Simoncelli.
\newblock Image quality assessment: Unifying structure and texture similarity.
\newblock \emph{TPAMI}, 44\penalty0 (5):\penalty0 2567--2581, 2020.

\bibitem[Dong et~al.(2014)Dong, Loy, He, and Tang]{srcnn}
Chao Dong, Chen~Change Loy, Kaiming He, and Xiaoou Tang.
\newblock Learning a deep convolutional network for image super-resolution.
\newblock In \emph{Computer Vision--ECCV 2014: 13th European Conference, Zurich, Switzerland, September 6-12, 2014, Proceedings, Part IV 13}, pages 184--199. Springer, 2014.

\bibitem[Esser et~al.(2021)Esser, Rombach, and Ommer]{vqgan}
Patrick Esser, Robin Rombach, and Bjorn Ommer.
\newblock Taming transformers for high-resolution image synthesis.
\newblock In \emph{Proceedings of the IEEE/CVF conference on computer vision and pattern recognition}, pages 12873--12883, 2021.

\bibitem[Goodfellow et~al.(2014)Goodfellow, Pouget-Abadie, Mirza, Xu, Warde-Farley, Ozair, Courville, and Bengio]{gan}
Ian Goodfellow, Jean Pouget-Abadie, Mehdi Mirza, Bing Xu, David Warde-Farley, Sherjil Ozair, Aaron Courville, and Yoshua Bengio.
\newblock Generative adversarial nets.
\newblock In \emph{NeurIPS}, 2014.

\bibitem[Heusel et~al.(2017)Heusel, Ramsauer, Unterthiner, Nessler, and Hochreiter]{fid}
Martin Heusel, Hubert Ramsauer, Thomas Unterthiner, Bernhard Nessler, and Sepp Hochreiter.
\newblock Gans trained by a two time-scale update rule converge to a local nash equilibrium.
\newblock \emph{NeurIPS}, pages 6626--6637, 2017.

\bibitem[Hu et~al.(2021)Hu, Shen, Wallis, Allen-Zhu, Li, Wang, Wang, and Chen]{lora}
Edward~J Hu, Yelong Shen, Phillip Wallis, Zeyuan Allen-Zhu, Yuanzhi Li, Shean Wang, Lu Wang, and Weizhu Chen.
\newblock Lora: Low-rank adaptation of large language models.
\newblock \emph{arXiv preprint arXiv:2106.09685}, 2021.

\bibitem[Karras et~al.(2019)Karras, Laine, and Aila]{ffhq}
Tero Karras, Samuli Laine, and Timo Aila.
\newblock A style-based generator architecture for generative adversarial networks.
\newblock In \emph{Proceedings of the IEEE/CVF conference on computer vision and pattern recognition}, pages 4401--4410, 2019.

\bibitem[Ke et~al.(2021)Ke, Wang, Wang, Milanfar, and Yang]{musiq}
Junjie Ke, Qifei Wang, Yilin Wang, Peyman Milanfar, and Feng Yang.
\newblock Musiq: Multi-scale image quality transformer.
\newblock In \emph{Proceedings of the IEEE/CVF International Conference on Computer Vision}, pages 5148--5157, 2021.

\bibitem[Ledig et~al.(2017)Ledig, Theis, Husz{\'a}r, Caballero, Cunningham, Acosta, Aitken, Tejani, Totz, Wang, et~al.]{srgan}
Christian Ledig, Lucas Theis, Ferenc Husz{\'a}r, Jose Caballero, Andrew Cunningham, Alejandro Acosta, Andrew Aitken, Alykhan Tejani, Johannes Totz, Zehan Wang, et~al.
\newblock Photo-realistic single image super-resolution using a generative adversarial network.
\newblock In \emph{Proceedings of the IEEE conference on computer vision and pattern recognition}, pages 4681--4690, 2017.

\bibitem[Li et~al.(2022)Li, Li, Xiong, and Hoi]{blip}
Junnan Li, Dongxu Li, Caiming Xiong, and Steven Hoi.
\newblock Blip: Bootstrapping language-image pre-training for unified vision-language understanding and generation.
\newblock In \emph{International Conference on Machine Learning}, pages 12888--12900. PMLR, 2022.

\bibitem[Li et~al.(2023)Li, Zhang, Liang, Cao, Liu, Gong, Zhang, Tang, Liu, Demandolx, et~al.]{lsdir}
Yawei Li, Kai Zhang, Jingyun Liang, Jiezhang Cao, Ce Liu, Rui Gong, Yulun Zhang, Hao Tang, Yun Liu, Denis Demandolx, et~al.
\newblock Lsdir: A large scale dataset for image restoration.
\newblock In \emph{Proceedings of the IEEE/CVF Conference on Computer Vision and Pattern Recognition}, pages 1775--1787, 2023.

\bibitem[Liang et~al.(2021)Liang, Cao, Sun, Zhang, Van~Gool, and Timofte]{swinir}
Jingyun Liang, Jiezhang Cao, Guolei Sun, Kai Zhang, Luc Van~Gool, and Radu Timofte.
\newblock Swinir: Image restoration using swin transformer.
\newblock In \emph{ICCVW}, pages 1833--1844, 2021.

\bibitem[Lim et~al.(2017{\natexlab{b}})Lim, Son, Kim, Nah, and Mu~Lee]{edsr}
Bee Lim, Sanghyun Son, Heewon Kim, Seungjun Nah, and Kyoung Mu~Lee.
\newblock Enhanced deep residual networks for single image super-resolution.
\newblock In \emph{Proceedings of the IEEE conference on computer vision and pattern recognition workshops}, pages 136--144, 2017{\natexlab{b}}.

\bibitem[Lin et~al.(2023)Lin, He, Chen, Lyu, Fei, Dai, Ouyang, Qiao, and Dong]{diffbir}
Xinqi Lin, Jingwen He, Ziyan Chen, Zhaoyang Lyu, Ben Fei, Bo Dai, Wanli Ouyang, Yu Qiao, and Chao Dong.
\newblock Diffbir: Towards blind image restoration with generative diffusion prior.
\newblock \emph{arXiv preprint arXiv:2308.15070}, 2023.

\bibitem[Liu et~al.(2024)Liu, Zhao, Zhuo, Lin, Qiao, Li, and Gao]{lumina-mgpt}
Dongyang Liu, Shitian Zhao, Le Zhuo, Weifeng Lin, Yu Qiao, Hongsheng Li, and Peng Gao.
\newblock Lumina-mgpt: Illuminate flexible photorealistic text-to-image generation with multimodal generative pretraining, 2024.

\bibitem[Liu et~al.(2023)Liu, Li, Wu, and Lee]{llava}
Haotian Liu, Chunyuan Li, Qingyang Wu, and Yong~Jae Lee.
\newblock Visual instruction tuning.
\newblock In \emph{NeurIPS}, 2023.

\bibitem[Mou et~al.(2024)Mou, Wang, Wu, Shan, and Zhang]{dem}
Chong Mou, Xintao Wang, Yanze Wu, Ying Shan, and Jian Zhang.
\newblock Empowering real-world image super-resolution with flexible interactive modulation.
\newblock \emph{IEEE Transactions on Pattern Analysis and Machine Intelligence}, 46\penalty0 (11):\penalty0 7317--7330, 2024.

\bibitem[Radford et~al.(2021)Radford, Kim, Hallacy, Ramesh, Goh, Agarwal, Sastry, Askell, Mishkin, Clark, et~al.]{clip}
Alec Radford, Jong~Wook Kim, Chris Hallacy, Aditya Ramesh, Gabriel Goh, Sandhini Agarwal, Girish Sastry, Amanda Askell, Pamela Mishkin, Jack Clark, et~al.
\newblock Learning transferable visual models from natural language supervision.
\newblock In \emph{International conference on machine learning}, pages 8748--8763. PMLR, 2021.

\bibitem[Rombach et~al.(2021)Rombach, Blattmann, Lorenz, Esser, and Ommer]{vae}
Robin Rombach, Andreas Blattmann, Dominik Lorenz, Patrick Esser, and Björn Ommer.
\newblock High-resolution image synthesis with latent diffusion models, 2021.

\bibitem[Shazeer(2020)]{swiglu}
Noam Shazeer.
\newblock Glu variants improve transformer, 2020.

\bibitem[Su et~al.(2023)Su, Lu, Pan, Murtadha, Wen, and Liu]{rope}
Jianlin Su, Yu Lu, Shengfeng Pan, Ahmed Murtadha, Bo Wen, and Yunfeng Liu.
\newblock Roformer: Enhanced transformer with rotary position embedding, 2023.

\bibitem[Sun et~al.(2024{\natexlab{a}})Sun, Wu, Ma, Liu, Yi, and Zhang]{pisa-sr}
Lingchen Sun, Rongyuan Wu, Zhiyuan Ma, Shuaizheng Liu, Qiaosi Yi, and Lei Zhang.
\newblock Pixel-level and semantic-level adjustable super-resolution: A dual-lora approach, 2024{\natexlab{a}}.

\bibitem[Sun et~al.(2024{\natexlab{b}})Sun, Jiang, Chen, Zhang, Peng, Luo, and Yuan]{llamagen}
Peize Sun, Yi Jiang, Shoufa Chen, Shilong Zhang, Bingyue Peng, Ping Luo, and Zehuan Yuan.
\newblock Autoregressive model beats diffusion: Llama for scalable image generation.
\newblock \emph{arXiv preprint arXiv:2406.06525}, 2024{\natexlab{b}}.

\bibitem[Team(2024)]{Chameleon}
Chameleon Team.
\newblock Chameleon: Mixed-modal early-fusion foundation models.
\newblock \emph{arXiv preprint arXiv:2405.09818}, 2024.

\bibitem[Touvron et~al.(2023)Touvron, Lavril, Izacard, Martinet, Lachaux, Lacroix, Rozière, Goyal, Hambro, Azhar, Rodriguez, Joulin, Grave, and Lample]{llama}
Hugo Touvron, Thibaut Lavril, Gautier Izacard, Xavier Martinet, Marie-Anne Lachaux, Timothée Lacroix, Baptiste Rozière, Naman Goyal, Eric Hambro, Faisal Azhar, Aurelien Rodriguez, Armand Joulin, Edouard Grave, and Guillaume Lample.
\newblock Llama: Open and efficient foundation language models, 2023.

\bibitem[Wang et~al.(2023{\natexlab{a}})Wang, Chan, and Loy]{clipiqa}
Jianyi Wang, Kelvin~CK Chan, and Chen~Change Loy.
\newblock Exploring clip for assessing the look and feel of images.
\newblock In \emph{Proceedings of the AAAI Conference on Artificial Intelligence}, pages 2555--2563, 2023{\natexlab{a}}.

\bibitem[Wang et~al.(2023{\natexlab{b}})Wang, Yue, Zhou, Chan, and Loy]{stablesr}
Jianyi Wang, Zongsheng Yue, Shangchen Zhou, Kelvin~CK Chan, and Chen~Change Loy.
\newblock Exploiting diffusion prior for real-world image super-resolution.
\newblock \emph{arXiv preprint arXiv:2305.07015}, 2023{\natexlab{b}}.

\bibitem[Wang et~al.(2018)Wang, Yu, Dong, and Loy]{ost}
Xintao Wang, Ke Yu, Chao Dong, and Chen~Change Loy.
\newblock Recovering realistic texture in image super-resolution by deep spatial feature transform.
\newblock In \emph{Proceedings of the IEEE conference on computer vision and pattern recognition}, pages 606--615, 2018.

\bibitem[Wang et~al.(2021)Wang, Xie, Dong, and Shan]{realesrgan}
Xintao Wang, Liangbin Xie, Chao Dong, and Ying Shan.
\newblock Real-esrgan: Training real-world blind super-resolution with pure synthetic data.
\newblock In \emph{ICCVW}, pages 1905--1914, 2021.

\bibitem[Wang et~al.(2024)Wang, Zhang, Luo, Sun, Cui, Wang, Zhang, Wang, Li, Yu, et~al.]{emu3}
Xinlong Wang, Xiaosong Zhang, Zhengxiong Luo, Quan Sun, Yufeng Cui, Jinsheng Wang, Fan Zhang, Yueze Wang, Zhen Li, Qiying Yu, et~al.
\newblock Emu3: Next-token prediction is all you need.
\newblock \emph{arXiv preprint arXiv:2409.18869}, 2024.

\bibitem[Wang et~al.(2004)Wang, Bovik, Sheikh, and Simoncelli]{ssim}
Zhou Wang, Alan~C Bovik, Hamid~R Sheikh, and Eero~P Simoncelli.
\newblock Image quality assessment: from error visibility to structural similarity.
\newblock \emph{IEEE transactions on image processing}, 13\penalty0 (4):\penalty0 600--612, 2004.

\bibitem[Wei et~al.(2020)Wei, Xie, Lu, Zhan, Ye, Zuo, and Lin]{drealsr}
Pengxu Wei, Ziwei Xie, Hannan Lu, Zongyuan Zhan, Qixiang Ye, Wangmeng Zuo, and Liang Lin.
\newblock Component divide-and-conquer for real-world image super-resolution.
\newblock In \emph{ECCV}, pages 101--117, 2020.

\bibitem[Wu et~al.(2024{\natexlab{a}})Wu, Sun, Ma, and Zhang]{osediff}
Rongyuan Wu, Lingchen Sun, Zhiyuan Ma, and Lei Zhang.
\newblock One-step effective diffusion network for real-world image super-resolution.
\newblock \emph{arXiv preprint arXiv:2406.08177}, 2024{\natexlab{a}}.

\bibitem[Wu et~al.(2024{\natexlab{b}})Wu, Yang, Sun, Zhang, Li, and Zhang]{seesr}
Rongyuan Wu, Tao Yang, Lingchen Sun, Zhengqiang Zhang, Shuai Li, and Lei Zhang.
\newblock Seesr: Towards semantics-aware real-world image super-resolution.
\newblock In \emph{CVPR}, 2024{\natexlab{b}}.

\bibitem[Xie et~al.(2024)Xie, Mao, Bai, Zhang, Wang, Lin, Gu, Chen, Yang, and Shou]{showo}
Jinheng Xie, Weijia Mao, Zechen Bai, David~Junhao Zhang, Weihao Wang, Kevin~Qinghong Lin, Yuchao Gu, Zhijie Chen, Zhenheng Yang, and Mike~Zheng Shou.
\newblock Show-o: One single transformer to unify multimodal understanding and generation.
\newblock \emph{arXiv preprint arXiv:2408.12528}, 2024.

\bibitem[Yang et~al.(2022)Yang, Wu, Shi, Lao, Gong, Cao, Wang, and Yang]{maniqa}
Sidi Yang, Tianhe Wu, Shuwei Shi, Shanshan Lao, Yuan Gong, Mingdeng Cao, Jiahao Wang, and Yujiu Yang.
\newblock Maniqa: Multi-dimension attention network for no-reference image quality assessment.
\newblock In \emph{CVPR}, pages 1191--1200, 2022.

\bibitem[Yang et~al.(2023)Yang, Wu, Ren, Xie, and Zhang]{pasd}
Tao Yang, Rongyuan Wu, Peiran Ren, Xuansong Xie, and Lei Zhang.
\newblock Pixel-aware stable diffusion for realistic image super-resolution and personalized stylization.
\newblock \emph{arXiv preprint arXiv:2308.14469}, 2023.

\bibitem[Yu et~al.(2024)Yu, Gu, Li, Hu, Kong, Wang, He, Qiao, and Dong]{supir}
Fanghua Yu, Jinjin Gu, Zheyuan Li, Jinfan Hu, Xiangtao Kong, Xintao Wang, Jingwen He, Yu Qiao, and Chao Dong.
\newblock Scaling up to excellence: Practicing model scaling for photo-realistic image restoration in the wild.
\newblock In \emph{CVPR}, 2024.

\bibitem[Zhang and Sennrich(2019)]{rmsnorm}
Biao Zhang and Rico Sennrich.
\newblock {Root Mean Square Layer Normalization}.
\newblock In \emph{Advances in Neural Information Processing Systems 32}, Vancouver, Canada, 2019.

\bibitem[Zhang et~al.(2015)Zhang, Zhang, and Bovik]{niqe}
Lin Zhang, Lei Zhang, and Alan~C Bovik.
\newblock A feature-enriched completely blind image quality evaluator.
\newblock \emph{TIP}, 24\penalty0 (8):\penalty0 2579--2591, 2015.

\bibitem[Zhang et~al.(2023{\natexlab{a}})Zhang, Rao, and Agrawala]{controlnet}
Lvmin Zhang, Anyi Rao, and Maneesh Agrawala.
\newblock Adding conditional control to text-to-image diffusion models.
\newblock In \emph{ICCV}, pages 3836--3847, 2023{\natexlab{a}}.

\bibitem[Zhang et~al.(2018{\natexlab{a}})Zhang, Isola, Efros, Shechtman, and Wang]{lpips}
Richard Zhang, Phillip Isola, Alexei~A Efros, Eli Shechtman, and Oliver Wang.
\newblock The unreasonable effectiveness of deep features as a perceptual metric.
\newblock In \emph{CVPR}, pages 586--595, 2018{\natexlab{a}}.

\bibitem[Zhang et~al.(2018{\natexlab{b}})Zhang, Li, Li, Wang, Zhong, and Fu]{rcan}
Yulun Zhang, Kunpeng Li, Kai Li, Lichen Wang, Bineng Zhong, and Yun Fu.
\newblock Image super-resolution using very deep residual channel attention networks.
\newblock In \emph{Proceedings of the European conference on computer vision (ECCV)}, pages 286--301, 2018{\natexlab{b}}.

\bibitem[Zhou et~al.(2024)Zhou, Yu, Babu, Tirumala, Yasunaga, Shamis, Kahn, Ma, Zettlemoyer, and Levy]{transfusion}
Chunting Zhou, Lili Yu, Arun Babu, Kushal Tirumala, Michihiro Yasunaga, Leonid Shamis, Jacob Kahn, Xuezhe Ma, Luke Zettlemoyer, and Omer Levy.
\newblock Transfusion: Predict the next token and diffuse images with one multi-modal model.
\newblock 2024.

\end{thebibliography}
}
\appendix
\onecolumn
\section{Appendix}
\label{sec:supp}
In the appendix, we first present more visual comparisons of the competing Real-ISR methods, then present the visual results in our ablation study. 

\subsection{More Visual Comparisons}
\label{sec:supp1}
\begin{figure*}[htbp]
  \centering
  \includegraphics[width=1.0\linewidth]{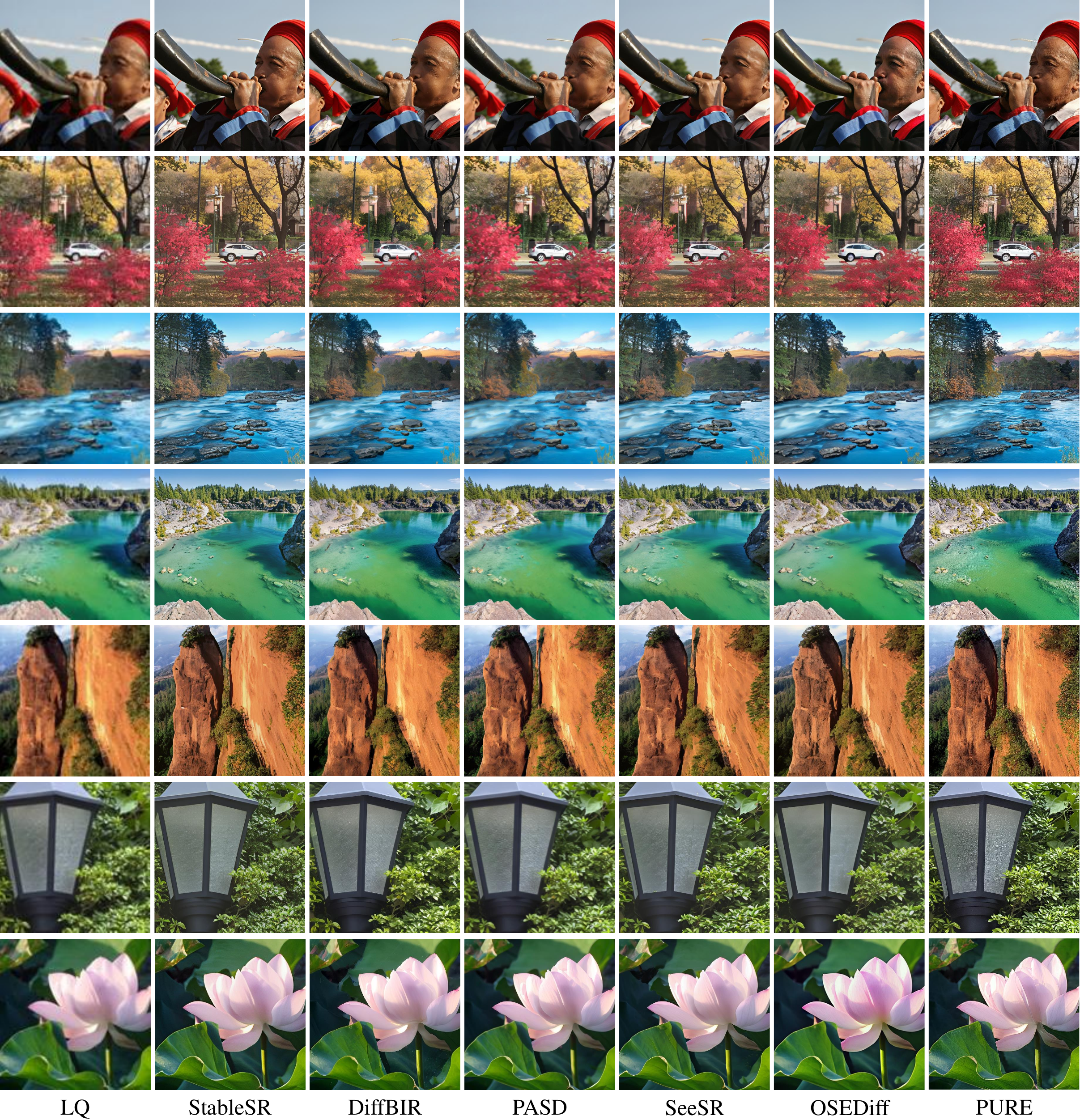}
  \caption{Visual comparisons of different Real-ISR methods on real-world images. Please zoom in for a better view.}
  \label{fig:supp1}
\end{figure*}

\begin{figure*}[htbp]
  \centering
  \includegraphics[width=1.0\linewidth]{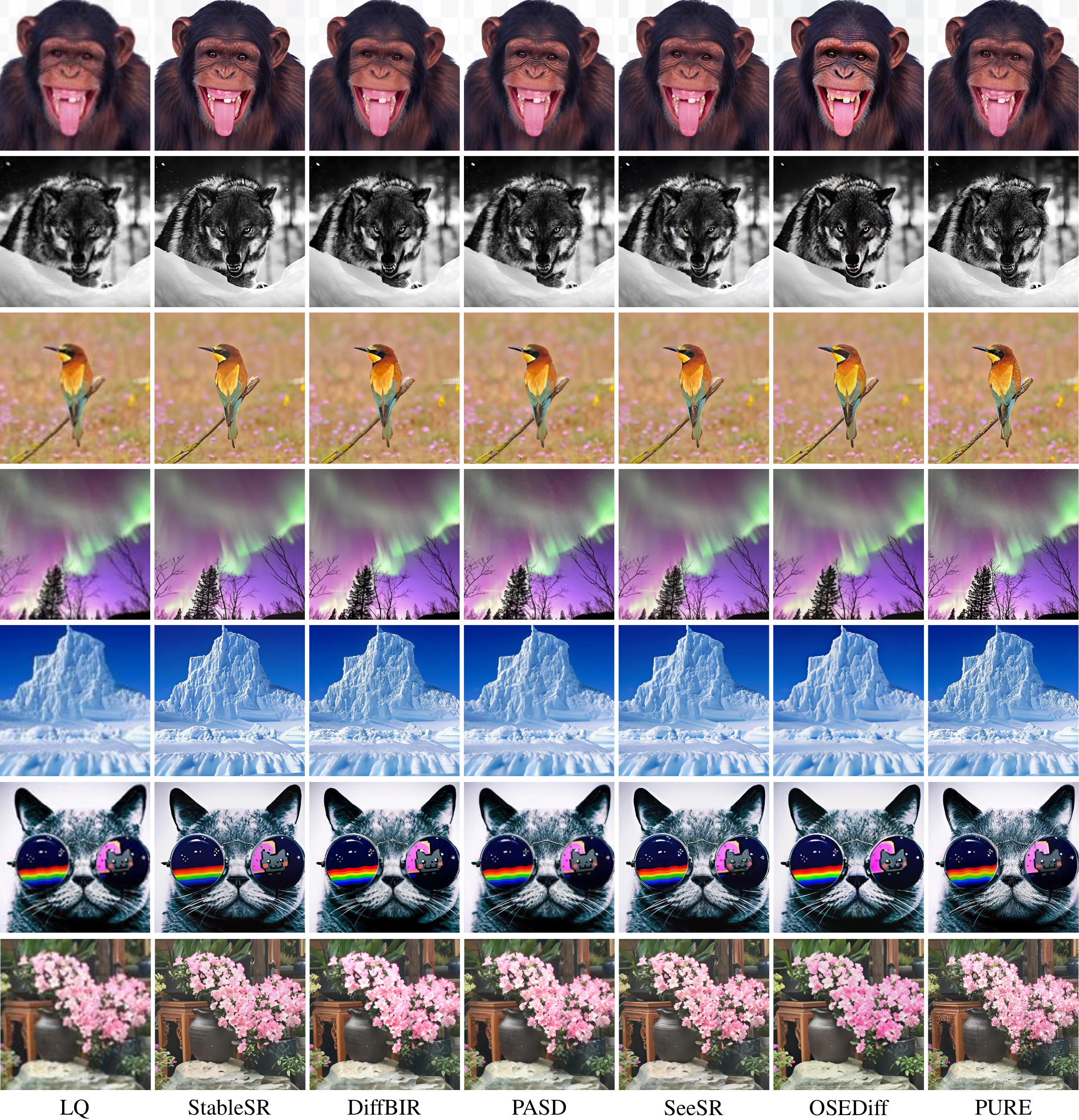}
  \caption{Visual comparisons of different Real-ISR methods on real-world images. Please zoom in for a better view.}
  \label{fig:supp2}
\end{figure*}

We present more visual comparisons with state-of-the-art diffusion-based Real-ISR methods on real-world images. As illustrated in ~\cref{fig:supp1} and ~\cref{fig:supp2}, our proposed PURE demonstrates superior restoration quality across diverse scenarios, including animals, plants, humans, and landscapes. The PURE framework not only effectively preserves the structural and textural integrity of LQ input images, but also generates more abundant and plausible details, achieving exceptional visual quality.

\subsection{Visual Results for Ablation Study}
\label{sec:supp2}
We provide visual comparison results for the ablation studies conducted in the main paper. 

\vspace{+2mm}

\begin{figure*}[htbp]
  \centering
  \includegraphics[width=0.8\linewidth]{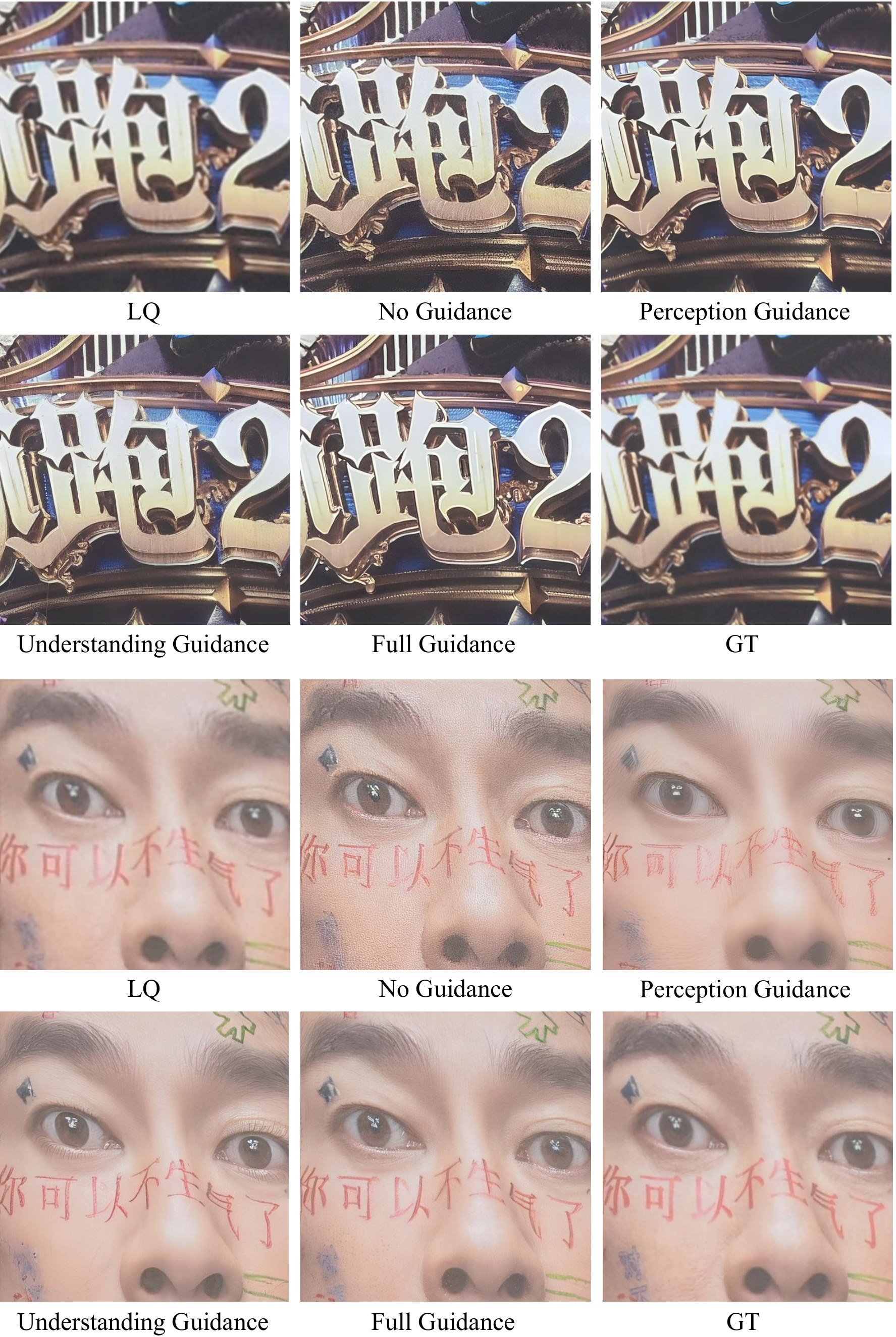}
  \caption{Visual comparisons for perception-understanding guidance ablation. Please zoom in for a better view.}
  \label{fig:supp_ablation_text}
\end{figure*}

\noindent \textbf{Perception-Understanding Guidance}. We first present visual comparisons on perception-understanding guidance, as illustrated in ~\cref{fig:supp_ablation_text}. The \textit{No Guidance} method demonstrates suboptimal restoration results with noticeable degradation in both cases, mainly due to its sole reliance on the LQ input image as the conditioning factor. Both \textit{Perception Guidance} and \textit{No Guidance} fail to generate rich details due to the absence of semantic description guidance. The \textit{Understanding Guidance} produces erroneous details (\eg, eyelashes), which can be attributed to the lack of adaptive degradation awareness. The \textit{Full Guidance} achieves the best restoration quality among all variants.

\begin{figure*}[htbp]
  \centering
  \includegraphics[width=0.8\linewidth]{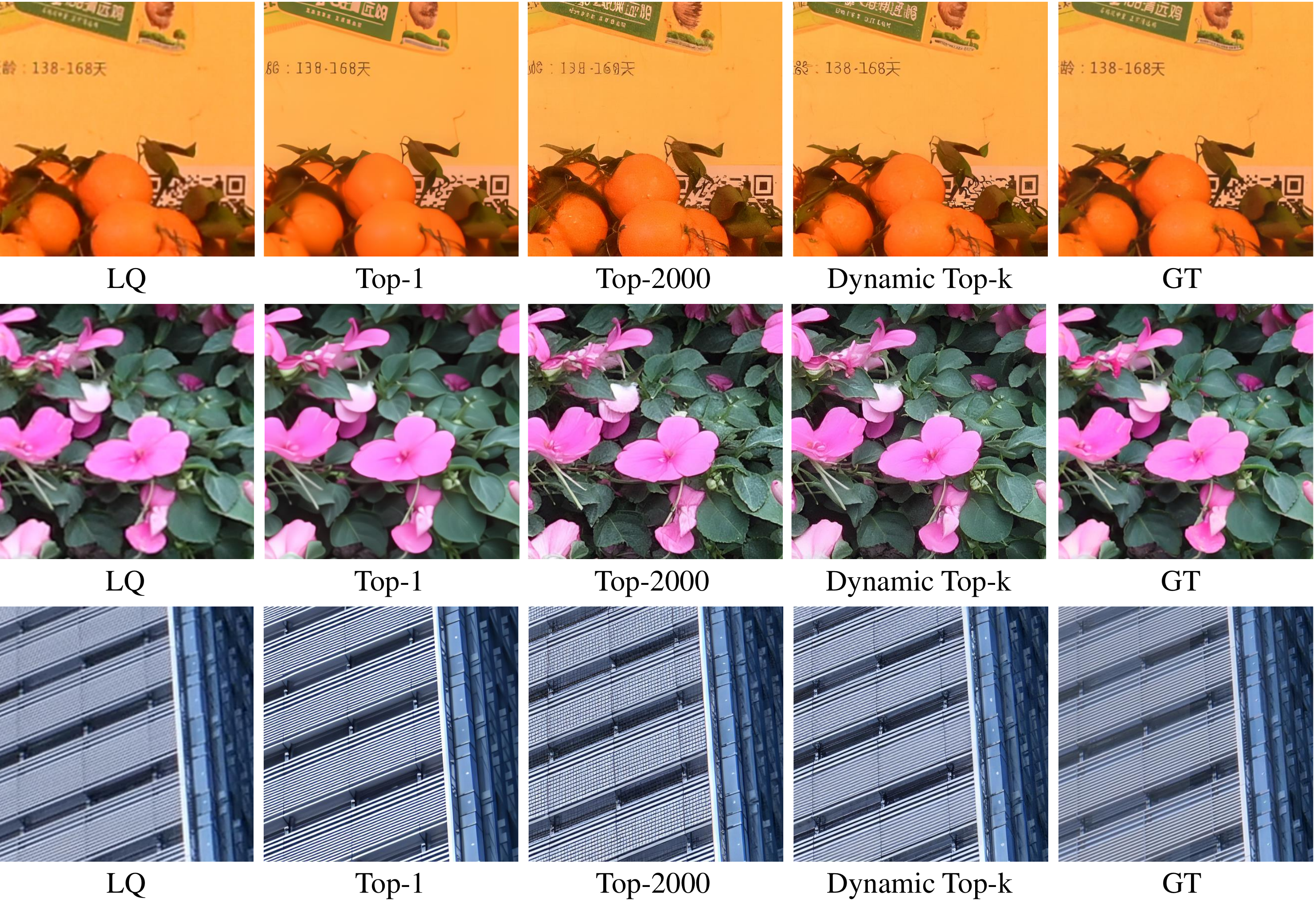}
  \caption{Visual comparisons for entropy-based Top-$k$ sampling ablation. Please zoom in for a better view.}
  \label{fig:supp_ablation_topk}
\end{figure*}

\vspace{+2mm}
\noindent \textbf{Entropy-based Top-$k$ Sampling}. To demonstrate the effectiveness of our proposed entropy-based Top-$k$ sampling strategy, we present visual comparison results by applying both fixed and dynamic Top-$k$ sampling approaches. As illustrated in ~\cref{fig:supp_ablation_topk}, the \textit{Top-1} strategy yields the smoothest yet most blurry restoration results, as its generative capability is constrained by selecting only the most probable image token at each step. In contrast, while the \textit{Top-2000} approach generates more details, it produces inferior visual perception quality with more erroneous textures. Notably, our entropy-based Top-$k$ sampling achieves the best restoration quality, maintaining superior fidelity while generating more realistic and authentic details.

\end{document}